\documentclass[runningheads]{llncs}

\usepackage{graphicx, tikz, comment, amsmath,amssymb, color}
\usepackage[accsupp]{axessibility}

\usepackage{enumitem, graphicx, multirow, amsfonts, amsmath, verbatim, algorithm, algorithmicx, algpseudocode, booktabs, arydshln, epsfig, amssymb, pifont, capt-of, wrapfig, lipsum, hyperref}

\begin{document}
\pagestyle{headings} \mainmatter

\def\ECCVSubNumber{6627} 
\title{Language-Driven Artistic Style Transfer}
\titlerunning{Fu et al. Language-Driven Artistic Style Transfer} 
\authorrunning{Fu, Wang, and Wang} 
\author{Tsu-Jui Fu$^\dagger$, Xin Eric Wang$^\ddagger$, William Yang Wang$^\dagger$}
\institute{$^\dagger$UC Santa Barbara~~~$^\ddagger$UC Santa Cruz
\\
{\tt \{tsu-juifu, william\}@cs.ucsb.edu}~~~{\tt xwang366@ucsc.edu}}

\maketitle

\begin{abstract}
Despite having promising results, style transfer, which requires preparing style images in advance, may result in lack of creativity and accessibility. Following human instruction, on the other hand, is the most natural way to perform artistic style transfer that can significantly improve controllability for visual effect applications. We introduce a new task---language-driven artistic style transfer (\texttt{LDAST})---to manipulate the style of a content image, guided by a text. We propose contrastive language visual artist (CLVA) that learns to extract visual semantics from style instructions and accomplish \texttt{LDAST} by the patch-wise style discriminator. The discriminator considers the correlation between language and patches of style images or transferred results to jointly embed style instructions. CLVA further compares contrastive pairs of content images and style instructions to improve the mutual relativeness. The results from the same content image can preserve consistent content structures. Besides, they should present analogous style patterns from style instructions that contain similar visual semantics. The experiments show that our CLVA is effective and achieves superb transferred results on \texttt{LDAST}.
\end{abstract}

\section{Introduction}
Style transfer~\cite{gatys2016nst,li2017wct,huang2017adain,park2019sa,li2019lst,jing2017review} adopts appearances and visual patterns from another reference style images to manipulate a content image. Artistic style transfer has a considerable application value for creative visual design, such as image stylization and video effect~\cite{somavarap2020st,zhang2019st,gao2018vst,huang2017vst}. However, it requires preparing collections of style image in advance. It even needs to redraw new references first if there is no expected style images, which is impractical due to an additional overhead. In contrast, language is the most natural way for humans to communicate. If a system can follow textual descriptions and automatically perform style transfer, we can significantly improve accessibility and controllability. 

\begin{figure}[!t]
\centering
    \includegraphics[width=\linewidth]{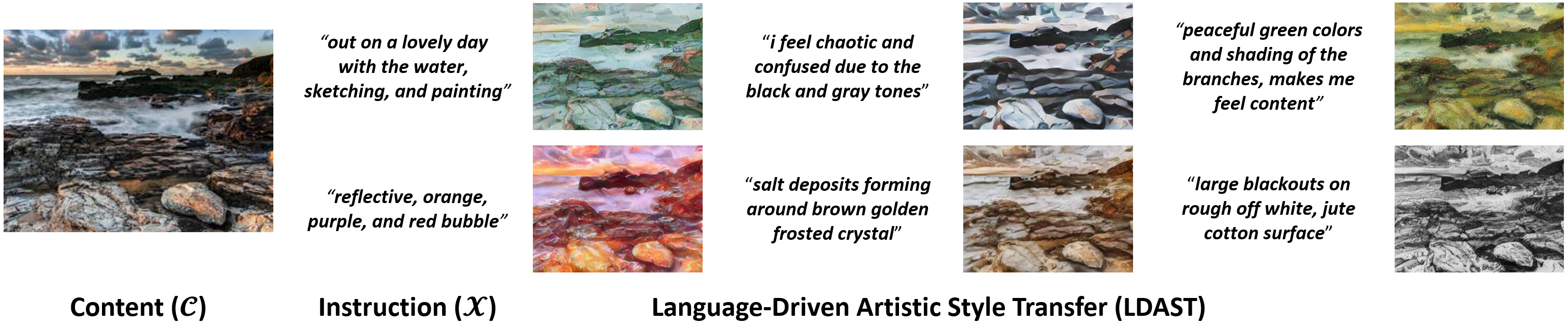}
    \vspace{-5ex}
    \caption{Language-driven Artistic Style Transfer (\texttt{LDAST}). \texttt{LDAST} performs style transfer for a content image $\mathcal{C}$, guided by the visual attribute (the lower row) or even the visual content and emotional effect (the upper row) from a style instruction $\mathcal{X}$.}
    \vspace{-3ex}
    \label{fig:ldast}
\end{figure}

In this paper, we introduce Language-driven Artistic Style Transfer (\texttt{LDAST}). As illustrated in Fig.~\ref{fig:ldast}, \texttt{LDAST} treats a content image and a text as the input, and the style transferred result is manipulated based on the style description. It should preserve the structure of the content yet simultaneously modifies the style pattern that corresponds to the instruction. \texttt{LDAST} is different from the general language-based image-editing (LBIE)~\cite{nam2018lbie,li2020lbie,liu2020open-edit,el-nouby2019ilbie} that aims at altering objects or properties of objects. The main challenge of \texttt{LDAST} is to extract visual semantics from language. Humans use not only explicit visual attributes but also visual content or emotional effects to describe style feelings. For example, it requires connecting \textit{``water, sketching, and painting''} or \textit{``peaceful, feel content''} with their visual concepts and further carrying out correlated style transfer.

We present contrastive language visual artist (CLVA), including language visual artist (LVA) and contrastive reasoning (CR), to perform style transfer conditioning on guided texts. LVA preserves content structures from content images $\mathcal{C}$ and extracts visual semantics from style instructions $\mathcal{X}$. LVA learns the latent style pattern based on the distinguishment between patches of style imags or transferred results from the patch-wise style discriminator. Furthermore, CR boosts by comparing contrastive pairs where relative content images or style instructions should present similar content structures or style patterns. 

To evaluate \texttt{LDAST}, we conduct experiments upon DTD$^2$~\cite{wu2020texture-text} and ArtEmis~\cite{achlioptas2021artemis}. DTD$^2$ provides texture images with its colors or texture patterns in text. ArtEmis collects explanations of visual contents and emotional effects for artworks. We treat these annotations as style instructions for the challenging \texttt{LDAST} concerning visual attributes or human style feelings. The experiments show that our CLVA is effective for \texttt{LDAST} and achieves superb yet efficient transferred results on both automatic metrics and human evaluation. Our contributions are four-fold: 
\begin{itemize}[topsep=0pt, noitemsep, leftmargin=*]
    \item We introduce \texttt{LDAST} that follows natural language for artistic style transfer;
    \item We present CLVA, which learns to extract explicit visual semantics from style instructions and provide sufficient style patterns for \texttt{LDAST};
    \item We conduct the evaluation on DTD$^2$ and ArtEmis to consider diverse style instructions with visual attributes and emotional effects;
    \item Extensive experiments and qualitative examples demonstrate that our CLVA outperforms baselines regarding both effectiveness and efficiency.
\end{itemize}

\section{Related Work}
\noindent \textbf{Artistic Style Transfer.}
Style transfer~\cite{gatys2016nst,jing2017review,wang2021rethink-st,chen2016st,gatys2017control-st,johnson2016ast,sanakoyeu2018ast} redraws an image with a specific style. Since being a popular form of art, incorporating painting with digital design can produce attractive visual effect (VFX). In general, style transfer can be divided into two categories: \textit{photorealistic} and \textit{artistic}. Photorealistic style transfer~\cite{luan2017pst,li2018photo-wct,yoo2019photo-wct2,park2020swap-ae} aims at applying reference styles on scenes without hurting details and satisfying contradictory objectives. By contrast, artistic style transfer~\cite{gatys2016nst,li2017wct,huang2017adain,park2019sa,li2019lst,liu2021ada-attn,chen2021dual-ast} captures style concepts from reference and modifies color distributions and texture patterns of content images. However, it requires preparing numerous style images in advance, which limits practicality of style transfer. To tackle this issue, \texttt{LDAST} allows following textual descriptions to perform \textit{artistic} style transfer and improves the accessibility of VFX design.

\vspace{1ex} \noindent \textbf{Language-based Image Editing.}
The general task of \texttt{LDAST} is language-based image editing (LBIE), which also uses language to edit input images. With rule-based instructions and predefined semantic labels, they~\cite{cheng13image-spirit,laput13pixel-tone} first carry out LBIE but under limited practicality. Inspired by text-to-image generation~\cite{reed2016t2i,zhang2017stack-gan,xu2018att-gan}, previous works~\cite{chen2018lbie,nam2018lbie,li2020lbie,xia2021tedi-gan,liu2020open-edit,el-nouby2019ilbie,fu2020sscr,fu2022m3l} perform LBIE by conditional GAN, which modifies the properties of objects in the image. In contrast, \texttt{LDAST} aims at preserving the scene structure from the content image and performing stylization guided by the style instruction.

\vspace{1ex} \noindent \textbf{CLIP-guided Optimization.}
Recently, based on the powerful visual-linguistic connection of CLIP~\cite{shi2020clip}, CLIP-guided image synthesis~\cite{aditya2021dall-e,nichol2021glide} has shown exciting results. StyleCLIP~\cite{patashnik2021style-clip} and NADA~\cite{gal2021nada} tweak the latent code of a pre-trained StyleGAN~\cite{kerras2020style-gan} for image editing. Since heavily relying on a pre-trained generator, both are confined to the training domain, and the results can only present limited stylization. CLIPstyler~\cite{kwon2022clip-styler} updates the style transfer network for target style patterns from the CLIP alignment. Though supporting arbitrary content images, CLIPstyler still requires hundreds of iterations and takes lots of time with considerable GPU memory, suffering from the efficiency and practicality overhead. Moreover, our experiments show that CLIP poorly captures detailed style patterns from instructions, which is intractable to perform explicit \texttt{LDAST}.

\section{Language-Driven Artistic Style Transfer}
\subsection{Overview of CLVA}
\begin{figure}[!t]
\centering
    \includegraphics[width=\linewidth]{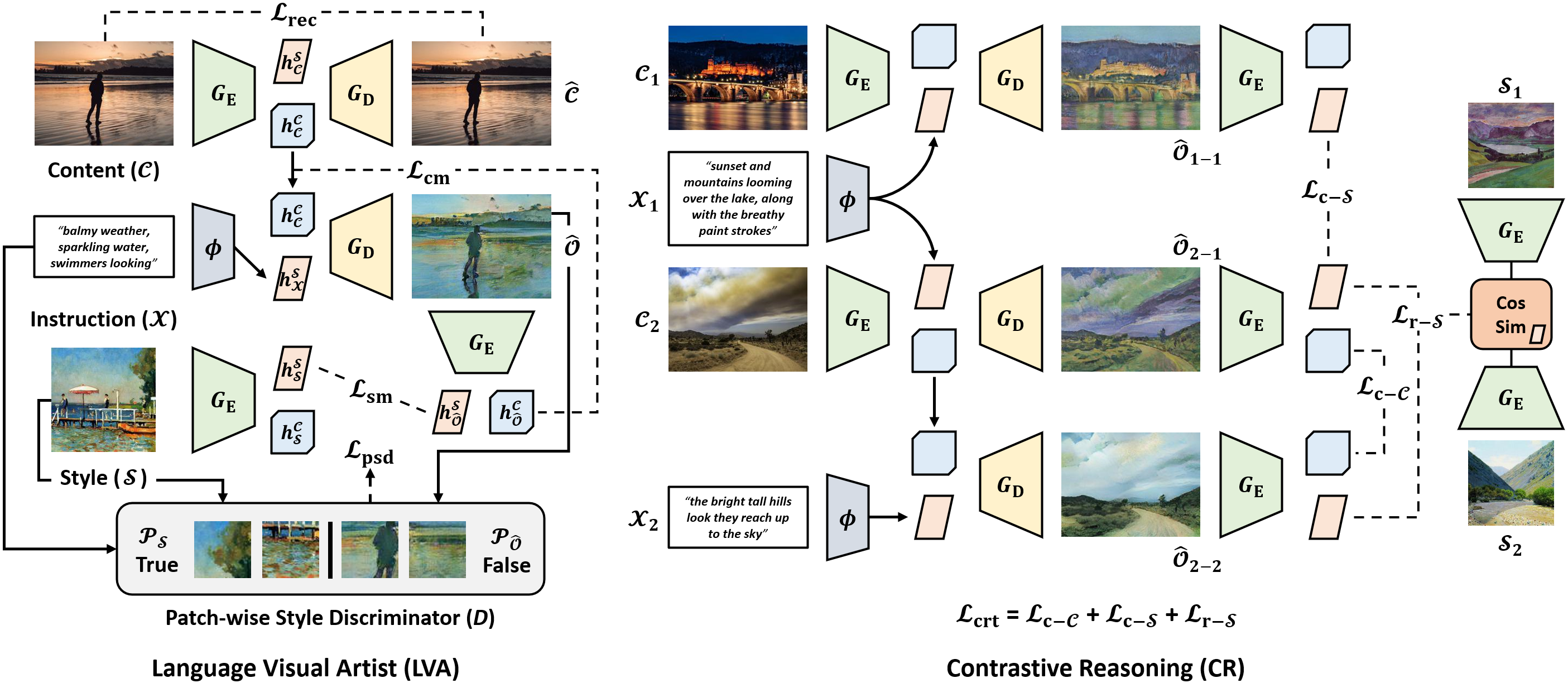}
    \vspace{-5ex}
    \caption{Contrastive language visual artist (CLVA). Language Visual Artist (LVA) learns to jointly embed style images $\mathcal{S}$ and style instructions $\mathcal{X}$ by the patch-wise style discriminator $D$ and perform \texttt{LDAST} for content images $\mathcal{C}$. Contrastive Reasoning (CR) compares contrasitve pairs to improve the relativeness between transferred results $\hat{\mathcal{O}}$.}
    \vspace{-3ex}
    \label{fig:clva}
\end{figure}

We introduce language-driven artistic style transfer (\texttt{LDAST}) to manipulate the style of a content image $\mathcal{C}$, guided by a style instruction $\mathcal{X}$, as illustrated in Fig.~\ref{fig:ldast}. For training, we have pairs of style images $\mathcal{S}$ with style instructions $\mathcal{X}$ to learn the mutual correlation. During testing, only $\mathcal{X}$ are provided for \texttt{LDAST} to carry out artistic style transfer purely relied on language. We present contrastive language visual artist (CLVA) in Fig.~\ref{fig:clva}. Language visual artist (LVA) extracts content structures from $\mathcal{C}$ and visual patterns from $\mathcal{X}$ to perform \texttt{LDAST}. LVA adopts the patch-wise style discriminator $D$ to connect extracted visual semantics to patches of paired style image ($\mathcal{P}_\mathcal{S}$ in Fig.~\ref{fig:clva}). Contrastive reasoning (CR) allows comparing contrastive pairs $\mathcal{C}_1$-$\mathcal{X}_1$, $\mathcal{C}_2$-$\mathcal{X}_1$, and $\mathcal{C}_2$-$\mathcal{X}_2$ of content image and style instruction. In this way, it should present consistent content structures from the same content image $\mathcal{C}_2$ or analogous style patterns from related style images $\mathcal{S}_1$ and $\mathcal{S}_2$, despite using different style instructions.

\subsection{Language Visual Artist (LVA)}
To tackle \texttt{LDAST}, language visual artist (LVA) first adopts visual encoder $G_\text{E}$ to extract the content feature $h^\mathcal{C}$ and the style feature $h^\mathcal{S}$ for an image. Text encoder $\phi$ also extracts the style instruction feature $h^\mathcal{S}_\mathcal{X}$ from an instruction. $h^\mathcal{C}$ is a spatial tensor containing the content structure feature, and $h^\mathcal{S}$ represents the global style pattern. $\mathcal{S}^S_\mathcal{X}$ embeds into the same space of $h^\mathcal{S}$ to reflect the extracted visual semantic. Then, visual decoder $G_\text{D}$ produces transferred results $\hat{\mathcal{O}}$ from $h^\mathcal{C}_\mathcal{C}$ and $h^\mathcal{S}_\mathcal{X}$, which performs style transfer by style instructions:
\begin{equation}
\begin{split}
    h^\mathcal{C}_\mathcal{C}, h^\mathcal{S}_\mathcal{C} &= G_\text{E}(\mathcal{C}), ~~~h^\mathcal{S}_\mathcal{X} = \phi (\mathcal{X}), \\
    \hat{\mathcal{O}} &= G_\text{D}(h^\mathcal{C}_\mathcal{C}, h^\mathcal{S}_\mathcal{X}).
\end{split}
\end{equation}
In particular, $G_\text{D}$ applies self-attention~\cite{zhang2019sa-gan,park2019sa} to fuse $h^\mathcal{C}$ and $h^\mathcal{S}$ over the global spatial dimension. There are two goals to train LVA for \texttt{LDAST}: ($i$) preserving \textit{content structures} from content images; ($ii$) presenting \textit{style patterns} correlated with visual semantics of style instructions.

\vspace{1ex} \noindent \textbf{Structure Reconstruction.}
To preserve content structures, we consider that visual decoder $G_\text{D}$ should be able to reconstruct input content images using extracted content features $h^\mathcal{C}_\mathcal{C}$ and style features $h^\mathcal{S}_\mathcal{C}$ from visual encoder $G_\text{E}$:
\begin{equation}
\begin{split}
\label{eq:rec}
    \hat{\mathcal{C}} &= G_\text{D}(h^\mathcal{C}_\mathcal{C}, h^\mathcal{S}_\mathcal{C}), \\
    \mathcal{L}_\text{rec} &= || \hat{\mathcal{C}} - \mathcal{C} ||_2,
\end{split}
\end{equation}
where the reconstruction loss $\mathcal{L}_\text{rec}$ is computed as the mean L2 difference between reconstructed content images $\hat{\mathcal{C}}$ and input content images $\mathcal{C}$.

\vspace{1ex} \noindent \textbf{Patch-wise Style Discriminator ($D$).}
Regarding style patterns, results $\hat{\mathcal{O}}$ guided by style instructions $\mathcal{X}$ are expected to present analogously to reference style images $\mathcal{S}$. To address the connection between linguistic from $\mathcal{X}$ and visual semantics from $\mathcal{S}$, we introduce the patch-wise style discriminator $D$. Inspired by texture synthesis~\cite{xian2018texture,gatys2015texture}, images with analogous patch patterns should appear perceptually similar texture patterns. $D$ tries to recognize the correspondence between an image patch $\mathcal{P}$ and a style instruction $\mathcal{X}$:
\begin{equation}
\begin{split}
\label{eq:psd}
    \mathcal{P}_{\hat{\mathcal{O}}}, \mathcal{P}_\mathcal{S} &= \text{Crop}(\hat{\mathcal{O}}), \text{Crop}(\mathcal{S}), \\
    \mathcal{L}_\text{psd} &= \log(1-D(\mathcal{P}_{\hat{\mathcal{O}}}, \mathcal{X})), \\
    \mathcal{L}_D &= \log(1-D(\mathcal{P}_{\hat{\mathcal{O}}}, \mathcal{X}))+\log(D(\mathcal{P}_\mathcal{S}, \mathcal{X})),
\end{split}
\end{equation}
where \texttt{Crop} is to randomly crop an image into patches. The patch-wise style loss $\mathcal{L}_\text{psd}$ aims at generating transferred results that are correlated with $\mathcal{X}$. Contrarily, by the discriminator loss $\mathcal{L}_D$, $D$ learns to distinguish that a patch $\mathcal{P}$ is from style images ($\mathcal{P}_\mathcal{S}$) or transferred results ($\mathcal{P}_{\hat{\mathcal{O}}}$). This adversarial loss~\cite{goodfellow2014gan,salehi2020gan} encourages that transferred results from style instructions are presented similarly with style images, which jointly embeds the extracted visual semantics.

\vspace{1ex} \noindent \textbf{Content Matching and Style Matching.}
To further enhance the alignment with inputs, inspired by cycle consistency~\cite{zhu2017cycle,lin2018cycle,qiao2019cycle,yi2017cycle}, we consider the content matching loss $\mathcal{L}_\text{cm}$ and the style matching loss $\mathcal{L}_\text{sm}$ of transferred results $\hat{\mathcal{O}}$. We adopt $G_\text{E}$ again to extract content features $h^\mathcal{C}_{\hat{\mathcal{O}}}$ and style features $h^\mathcal{S}_{\hat{\mathcal{O}}}$ for $\hat{\mathcal{O}}$, where $h^\mathcal{C}_{\hat{\mathcal{O}}}$ and $h^\mathcal{S}_{\hat{\mathcal{O}}}$ should correlate with $h^\mathcal{C}_\mathcal{C}$ from $\mathcal{C}$ and $h^\mathcal{S}_\mathcal{S}$ from $\mathcal{S}$:
\begin{equation}
\begin{split}
\label{eq:cm}
    (h^\mathcal{C}_{\hat{\mathcal{O}}}, h^\mathcal{S}_{\hat{\mathcal{O}}}), (h^\mathcal{C}_\mathcal{S}, h^\mathcal{S}_\mathcal{S}) &= G_\text{E}(\hat{\mathcal{O}}), G_\text{E}(S), \\
    \mathcal{L}_\text{cm}, \mathcal{L}_\text{sm} &= || h^\mathcal{C}_{\hat{\mathcal{O}}} - h^\mathcal{C}_\mathcal{C} ||_2, || h^\mathcal{S}_{\hat{\mathcal{O}}} - h^\mathcal{S}_\mathcal{S} ||_2.
\end{split}
\end{equation}
Therefore, transferred results are required to align with content structures and style patterns from inputs, which meets the goal of \texttt{LDAST}.

\subsection{Contrastive Reasoning (CR)}
The content image should transfer to various styles while preserving the same structure. Related style instructions can apply analogous style patterns to arbitrary content images. As shown in Fig.~\ref{fig:clva}, contrastive reasoning (CR) compares content structures or style patterns from transferred results of contrastive pairs. The contrastive pair consists of two different content images $\mathcal{C}_1$ and $\mathcal{C}_2$ with two reference styles $\{\mathcal{S}_1, \mathcal{X}_1\}$ and $\{\mathcal{S}_2, \mathcal{X}_2\}$. We follow the LVA inference to acquire cross results for pairs of content images and style instructions:
\begin{align}
    (h^\mathcal{C}_{\mathcal{C}_1}, h^\mathcal{S}_{\mathcal{C}_1}), (h^\mathcal{C}_{\mathcal{C}_2}, h^\mathcal{S}_{\mathcal{C}_2}) &= G_\text{E}(\mathcal{C}_1), G_\text{E}(\mathcal{C}_2), \\
    h^\mathcal{S}_{\mathcal{X}_1}, h^\mathcal{S}_{\mathcal{X}_2} &= \phi (\mathcal{X}_1), \phi (\mathcal{X}_2), \nonumber \\
    \hat{\mathcal{O}}_{\mathcal{C}_1\text{-}\mathcal{X}_1}, \hat{\mathcal{O}}_{\mathcal{C}_1\text{-}\mathcal{X}_2} &= G_\text{D}(h^\mathcal{C}_{\mathcal{C}_1}, h^\mathcal{S}_{\mathcal{X}_1}), G_\text{D}(h^\mathcal{C}_{\mathcal{C}_1}, h^\mathcal{S}_{\mathcal{X}_2}), \nonumber \\
    \hat{\mathcal{O}}_{\mathcal{C}_2\text{-}\mathcal{X}_1}, \hat{\mathcal{O}}_{\mathcal{C}_2\text{-}\mathcal{X}_2}  &= G_\text{D}(h^\mathcal{C}_{\mathcal{C}_2}, h^\mathcal{S}_{\mathcal{X}_1}), G_\text{D}(h^\mathcal{C}_{\mathcal{C}_2}, h^\mathcal{S}_{\mathcal{X}_2}). \nonumber
\end{align}

\vspace{1ex} \noindent \textbf{Consistent Matching.}
Transferred results should present similar content structures ($\hat{\mathcal{O}}_{\mathcal{C}_2\text{-}\mathcal{X}_1}$ and $\hat{\mathcal{O}}_{\mathcal{C}_2\text{-}\mathcal{X}_2}$) or analogous style patterns ($\hat{\mathcal{O}}_{\mathcal{C}_1\text{-}\mathcal{X}_1}$ and $\hat{\mathcal{O}}_{\mathcal{C}_2\text{-}\mathcal{X}_1}$) if using the same content image ($\mathcal{C}_2$) or the same style instruction ($\mathcal{X}_1$):
\begin{align}
    h^\mathcal{C}_{\hat{\mathcal{O}}_{\mathcal{C}_i\text{-}\mathcal{X}_j}} &= G_\text{E}(\hat{\mathcal{O}}_{\mathcal{C}_i\text{-}\mathcal{X}_j}), \\
    \mathcal{L}_{\text{c}-\mathcal{C}} = || h^\mathcal{C}_{\hat{\mathcal{O}}_{\mathcal{C}_1\text{-}\mathcal{X}_1}} &- h^\mathcal{C}_{\hat{\mathcal{O}}_{\mathcal{C}_1\text{-}\mathcal{X}_2}} ||_2 + || h^\mathcal{C}_{\hat{\mathcal{O}}_{\mathcal{C}_2\text{-}\mathcal{X}_1}} - h^\mathcal{C}_{\hat{\mathcal{O}}_{\mathcal{C}_2\text{-}\mathcal{X}_2}} ||_2, \nonumber \\
    \mathcal{L}_{\text{c}-\mathcal{S}} = || h^\mathcal{S}_{\hat{\mathcal{O}}_{\mathcal{C}_1\text{-}\mathcal{X}_1}} &- h^\mathcal{S}_{\hat{\mathcal{S}}_{2-1}} ||_2 + || h^\mathcal{S}_{\hat{\mathcal{O}}_{\mathcal{C}_1\text{-}\mathcal{X}_2}} - h^\mathcal{S}_{\hat{\mathcal{O}}_{\mathcal{C}_2\text{-}\mathcal{X}_2}} ||_2, \nonumber
\end{align}
where \textit{consistent matching} of content structure $\mathcal{L}_{\text{c}-\mathcal{C}}$ or style pattern $\mathcal{L}_{\text{c}-\mathcal{S}}$ is aligned by content features or style features, extracted by $G_\text{E}$.

\begin{algorithm}[t]
    \begin{algorithmic}[1]
    \scriptsize
        \State $G_\text{E}$, $G_\text{D}$: Visual Encoder, Visual Decoder
        \State $\phi$: Text Encoder
        \State $D$: Patch-wise Style Discriminator
        \While{TRAIN\_VLA}
            \State $\mathcal{C}$, $\{\mathcal{S}, \mathcal{X}\}$ $\gets$ Sampled content/style
            \\
            \State $h^\mathcal{C}_\mathcal{C}$, $h^\mathcal{S}_\mathcal{C}$ $\gets$ $G_\text{E}(\mathcal{C})$~~~~~$\hat{\mathcal{C}}$ $\gets$ $G_\text{D}(h^\mathcal{C}_\mathcal{C},h^\mathcal{S}_\mathcal{C})$
            \State $\mathcal{L}_\text{rec}$ $\gets$ Reconstruction loss \Comment{Eq.~\ref{eq:rec}}
            \State $h^\mathcal{S}_\mathcal{X}$ $\gets$ $\phi(\mathcal{X})$~~~~~$\hat{\mathcal{O}}$ $\gets$ $G_\text{D}(h^\mathcal{C}_\mathcal{C},h^\mathcal{S}_\mathcal{X})$
            \State $\mathcal{P}_\mathcal{S}$, $\mathcal{P}_{\hat{\mathcal{O}}}$ $\gets$ $\text{Crop}(\mathcal{S})$, $\text{Crop}(\hat{\mathcal{O}})$
            \State $\mathcal{L}_\text{psd}$ $\gets$ Patch-wise style loss \Comment{Eq.~\ref{eq:psd}}
            \State ($h^\mathcal{C}_{\hat{\mathcal{O}}}$, $h^\mathcal{S}_{\hat{\mathcal{O}}}$), ( $h^\mathcal{C}_\mathcal{S}$, $h^\mathcal{S}_\mathcal{S}$) $\gets$ $G_\text{E}(\hat{\mathcal{O}})$, $G_\text{E}(\mathcal{S})$
            \State $\mathcal{L}_\text{cm}$ $\gets$ Content matching loss \Comment{Eq.~\ref{eq:cm}}
            \State $\mathcal{L}_\text{sm}$ $\gets$ Style matching loss \Comment{Eq.~\ref{eq:cm}}
            \\
            \State $\mathcal{L}_G$ $\gets$ $\mathcal{L}_\text{rec}+\mathcal{L}_\text{psd}+\mathcal{L}_\text{cm}+\mathcal{L}_\text{sm}$
            \State $\mathcal{L}_D$ $\gets$ Discriminator loss for D \Comment{Eq.~\ref{eq:psd}}
            \State Update $G_\text{E}$, $G_\text{D}$, $\phi$ by minimizing $\mathcal{L}_G$
            \State Update $D$ by maximizing $\mathcal{L}_D$
        \EndWhile
    \end{algorithmic}
    \caption{Training Process of Language Visual Artist (LVA)}
    \label{algo:lva}
\end{algorithm}

\vspace{1ex} \noindent \textbf{Relative Matching.}
Apart from consistent matching, distinct style instructions, which imply corresponding visual semantics, should still present relative patterns. For example, we can only discover \textit{``reach up to the sky''} literally from $\mathcal{X}_2$. If comparing reference style images $\mathcal{S}_1$ and $\mathcal{S}_2$, we can perceive the sharing of a similar style pattern and link the visual concept of \textit{``bright tall hills''} in $\mathcal{X}_2$ to \textit{``mountains looming over the lake''} in $\mathcal{X}_1$. We define \textit{relative matching} $\mathcal{L}_{\text{r}-\mathcal{S}}$ with the cosine similarity (CosSim) between reference style images:
\begin{equation}
\begin{split}
    (h^\mathcal{C}_{\mathcal{S}_i}, h^\mathcal{S}_{\mathcal{S}_i}) &= G_\text{E}(\mathcal{S}_i), \\
    r &= \text{CosSim}(h^\mathcal{S}_{\mathcal{S}_1}, h^\mathcal{S}_{\mathcal{S}_2}), \\
    \mathcal{L}_{\text{r}-\mathcal{S}} = ( & || h^\mathcal{S}_{\hat{\mathcal{O}}_{\mathcal{C}_1\text{-}\mathcal{X}_1}} - h^\mathcal{S}_{\hat{\mathcal{O}}_{\mathcal{C}_1\text{-}\mathcal{X}_2}} ||_2 + \\ 
    & || h^\mathcal{S}_{\hat{\mathcal{O}}_{\mathcal{C}_2\text{-}\mathcal{X}_1}} - h^\mathcal{S}_{\hat{\mathcal{O}}_{\mathcal{C}_2\text{-}\mathcal{X}_2}} ||_2 ) \cdot r.
\end{split}
\end{equation}
When style images are related, it has to align style features to certain extent even if paired style instructions are different. Otherwise, $\mathcal{L}_{\text{r}-\mathcal{S}}$ will be close to $0$ and ignore this unrelated style pair. The overall contrastive reasoning loss $\mathcal{L}_\text{ctr}$ considers both consistent matching and relative matching:
\begin{equation}
\label{eq:ctr}
    \mathcal{L}_\text{ctr} = \mathcal{L}_{\text{c}-\mathcal{C}}+\mathcal{L}_{\text{c}-\mathcal{S}}+\mathcal{L}_{\text{r}-\mathcal{S}}.
\end{equation}

\subsection{Learning of CLVA}
For each epoch of CLVA training, we first train with the LVA process and then CR. As algo.~\ref{algo:lva}, we consider reconstruction loss $\mathcal{L}_\text{rec}$ to preserve content structure and patch-wise style loss $\mathcal{L}_\text{psd}$ between style instruction and visual pattern of transferred results. Both content matching loss $\mathcal{L}_\text{cm}$ and style matching loss $\mathcal{L}_\text{sm}$ enhance the matching with the inputs. Simultaneously, we update $D$ by maximizing discriminator loss $\mathcal{L}_D$ to distinguish between true patches $\mathcal{P}_\mathcal{S}$ or false patches $\mathcal{P}_{\hat{\mathcal{O}}}$, concerning style instructions. During CR, contrastive pairs of content images and style instructions are randomly sampled, and the transferred results are across produced. We further update by minimizing contrastive reasoning loss $\mathcal{L}_\text{ctr}$ to allow considering content consistency and mutual style relativeness. The overall optimization of CLVA is summarized as:
\begin{equation}
\begin{split}
    \mathcal{L}_G = & \mathcal{L}_\text{rec}+\mathcal{L}_\text{psd}+\mathcal{L}_\text{cm}+\mathcal{L}_\text{sm}, \\
    \min_{G, \phi} \max_{D} ~& \mathcal{L}_G+\mathcal{L}_D+\mathcal{L}_\text{ctr}.
\end{split}
\end{equation}

\section{Experiments}
\subsection{Experimental Setup}
\noindent \textbf{Dataset.}
To evaluate our CLVA, we consider DTD$^2$~\cite{cimpoi14dtd} and ArtEmis~\cite{achlioptas2021artemis} as reference style instructions. DTD$^2$ contains 5K texture images with its natural descriptions for visual attributes such as colors and texture patterns. ArtEmis provides 80K artworks from WikiArt\footnote{WikiArt:~\url{https://www.wikiart.org}} with annotations of visual contents and emotional effects as human style feelings. We also collect 15K wallpapers from WallpapersCraft\footnote{WallpapersCraft: \url{https://wallpaperscraft.com/}}, which presents diverse scenes as content images. Each content image is resized to 256x192 in our experiment. We randomly sample 100 unseen content images and 100 testing reference styles to evaluate the generalizability of \texttt{LDAST}. Note that both style images and style instructions appear for training, but only style instructions are accessible during testing.

\vspace{1ex} \noindent \textbf{Evaluation Metrics.}
To support large-scale evaluation, we treat transferred results directly from style images as semi-groundtruth (Semi-GT)~\cite{al-sarraf2014semi-gt,borkar2010semi-gt,salvo2013semi-gt} by the SOTA style transfer AdaAttn~\cite{liu2021ada-attn}. We apply the following metrics:
\begin{itemize}[topsep=0pt, noitemsep, leftmargin=*]
    \item \textbf{SSIM}~\cite{wang2004ssim} compares images in the luminance, contrast, and structure aspects. A higher SSIM has a higher structural similarity;
    \item \textbf{Percept}~\cite{johnson2016ast} computes from the gram matrix of visual features. A lower Percept loss shows that two images share a similar style pattern;
    \item \textbf{FAD}~\cite{heusel2017fid} is computed by the mean L2 distance of the activations from the InceptionV3~\cite{szegedy2016incv3} feature. As a distance metric, a lower FAD represents that \texttt{LDAST} results and Semi-GT are more relevant.
\end{itemize}
Note that we consider SSIM and FAD to compare with Semi-GT and calculate Percept loss directly with reference style images. Apart from visual similarity, we consider the correlation between style instructions and \texttt{LDAST} results:
\begin{itemize}[topsep=0pt, noitemsep, leftmargin=*]
    \item \textbf{VLS}~\cite{wu2021godiva} calculates the cosine similarity between each other from CLIP~\cite{shi2020clip}.
\end{itemize}
Since each metric has different deficiencies, we also conduct a comprehensive human evaluation from aspects of content, instruction, and style matching. We randomly sample 75 \texttt{LDAST} results and adopt MTurk\footnote{Amazon Mechanical Turk: \url{https://www.mturk.com}} to rank over all methods. We also hire 3 MTurkers for each task to avoid the potential ranking bias.

\vspace{1ex} \noindent \textbf{Baselines.}
We conduct baselines for LDAST from various aspects:
\begin{itemize}[topsep=0pt, noitemsep, leftmargin=*]
    \item \textbf{Style Transfer}: We consider previous artistic style transfer methods NST~\cite{gatys2016nst}, WCT~\cite{li2017wct}, AdaIn~\cite{huang2017adain}, SANet~\cite{park2019sa}, and LST~\cite{li2019lst} that support arbitrary contetn images. We use the same style (instruction and image) encoding from our CLVA as style features and follow their own training process to perform \texttt{LDAST} upon them. Due to the space issue, we only show the comparison with more recent SANet and LST. Please refer to Appendix for the complete results. 
    \item \textbf{Language-based Image Editing}: We adopt ManiGAN~\cite{li2020lbie} with affine combination module (ACM) as the general language-based editing baseline, where it modifies the content image by the style instruction. We treat normal style transferred results as groundtruth for ManiGAN to learn from.
    \item \textbf{CLIP-based Optimization}: StyleCLIP~\cite{patashnik2021style-clip}, NADA~\cite{gal2021nada}, and CLIPstyler~\cite{kwon2022clip-styler} manipulate the content image based on the CLIP alignment of the guided instruction. Since StyleCLIP and NADA are restricted by the pre-trained generator, we compare them with the training domains of car and church. Differently, CLIPstyler can carry out arbitrary content images for \texttt{LDAST}.
\end{itemize}

%\vspace{1ex} \noindent \textbf{Implementation Detail.}
%We adopt VGG-19~\cite{simonyan2015vgg,park2019sa} as our visual encoder $G_\text{E}$ and visual decoder $G_\text{D}$. Text encoder $\phi$ first adopts RoBERTa~\cite{liu2019roberta,reimers2019roberta} for a general linguistic and then expands its spatial dimension to jointly embed with style features. We follow the self-attention layer from SANet~\cite{park2019sa} to fuse between content and style features in $G_\text{D}$. The patch-wise style discriminator $D$ contains a similar architecture with a dense layer to determine the correlation between instructions and image patches. Both $G_\text{E}$ and $G_\text{D}$ are initialized from SANet and further update during the CLVA training process. We adopt Adam~\cite{kingma2015adam} to optimize CLVA with learning rate 3e-4 for $\mathcal{L}_G$, 1e-4 for $\mathcal{L}_D$, and 3e-5 for $\mathcal{L}_\text{ctr}$.

\subsection{Quantitative Results}
\noindent \textbf{Instruction with Visual Attributes.}
Table~\ref{table:dtd} illustrates the comparison of \texttt{LDAST} with baselines on DTD$^2$. As regards automatic metrics, CLVA preserves content structures (highest 36.65 SSIM) and stylizes with related visual attributes to style images (lowest 0.2033 Percept loss). Furthermore, CLVA brings out the highest overall similarity as Semi-GT (lowest 0.1493 FAD). Since CLIPstyler directly optimizes by CLIP~\cite{shi2020clip}, it makes the highest VLS. Through the patch-wise discriminator, our CLVA can still produce style patterns correlated to given instructions (competitive 24.00 VLS) even without the pre-trained CLIP.

The human evaluation investigates the matching between transferred results with content images (Content), style instructions (Instruction), style images (Style), and Semi-GT (Semi-GT). In particular, content and instruction matching are the two most crucial, which concern the goal of \texttt{LDAST}: \textit{content structure preservation} and \textit{style pattern presentation}; style image and semi-gt matching are provided for different comparing targets from a human aspect. The results are calculated by the mean ranking score (from 1 to 5, the higher is better) of each method. In general, MTurkers indicate that our CLVA has an apparent advantage in preserving content structures (highest 3.852 Content) and presenting aligned style patterns (highest 3.742 Instruction). Though with the aid of CLIP, CLIPstyler is still behind CLVA (-0.4 Instruction), with an even higher gap in style image matching (-0.5 Style). Contributed by contrastive reasoning that compares the mutual relativeness between pairs of contents and instructions, CLVA can stylize with the captured visual attributes. We adopt Pearson correlation and investigate the coefficients between automatic metrics and human evaluation as 77.2 (FAD→Instruction), 84.5 (FAD→Semi-GT), 81.3 (VLS→Instruction), and 77.8 (VLS→Semi-GT). This high correlation indicates that our metric design is adequate for evaluating large-scale LDAST experiments. The even higher 88.2 correlation (Instruction→Semi-GT) between instruction and Semi-GT matching in human evaluation further supports the usage of Semi-GT.

\begin{table}[t]
\centering \scriptsize
    \begin{tabular}[t]{cccccccccc}
        \toprule
        ~ & \multicolumn{4}{c}{\textbf{Automatic Metrics}} & ~ & \multicolumn{4}{c}{\textbf{Human Evaluation}} \\
        \cmidrule{2-5} \cmidrule{7-10} 
        Method & SSIM$\uparrow$ & Percept$\downarrow$ & FAD$\downarrow$ & VLS$\uparrow$ & ~ & Content$\uparrow$ & Instruction$\uparrow$ & Style$\uparrow$ & Semi-GT$\uparrow$ \\
        \midrule
        SANet~\cite{park2019sa} & 35.50 & \underline{0.2129} & 0.1627 & 23.57 & ~ & 2.701 & 2.477 & 2.738 & 2.630 \\
        LST~\cite{li2019lst} & \underline{34.84} & 0.2129 & \underline{0.1533} & 23.16 & ~ & 2.743 & 2.831 & 2.651 & 2.528 \\
        ManiGAN~\cite{li2020lbie} & 32.70 & 0.2401 & 0.1663 & 23.25 & ~ & 2.757 & 2.562 & 2.937 & 2.922 \\
        CLIPstyler~\cite{kwon2022clip-styler} & 25.24 & 0.2598 & 0.1818 & \textbf{24.62} & ~ & \underline{2.948} & \underline{3.388} & \underline{3.073} & \underline{3.265} \\
        CLVA & \textbf{36.65} & \textbf{0.2033} & \textbf{0.1493} & \underline{24.00} & ~ & \textbf{3.852} & \textbf{3.742} & \textbf{3.603} & \textbf{3.655} \\
        \bottomrule
        \\
    \end{tabular}
    \vspace{-1ex}
    \caption{Testing results of \texttt{LDAST} using visual attribute instructions on DTD$^2$.}
    \vspace{-4ex}
    \label{table:dtd}
\end{table}
\begin{figure}[!t]
\centering
    \includegraphics[width=\linewidth]{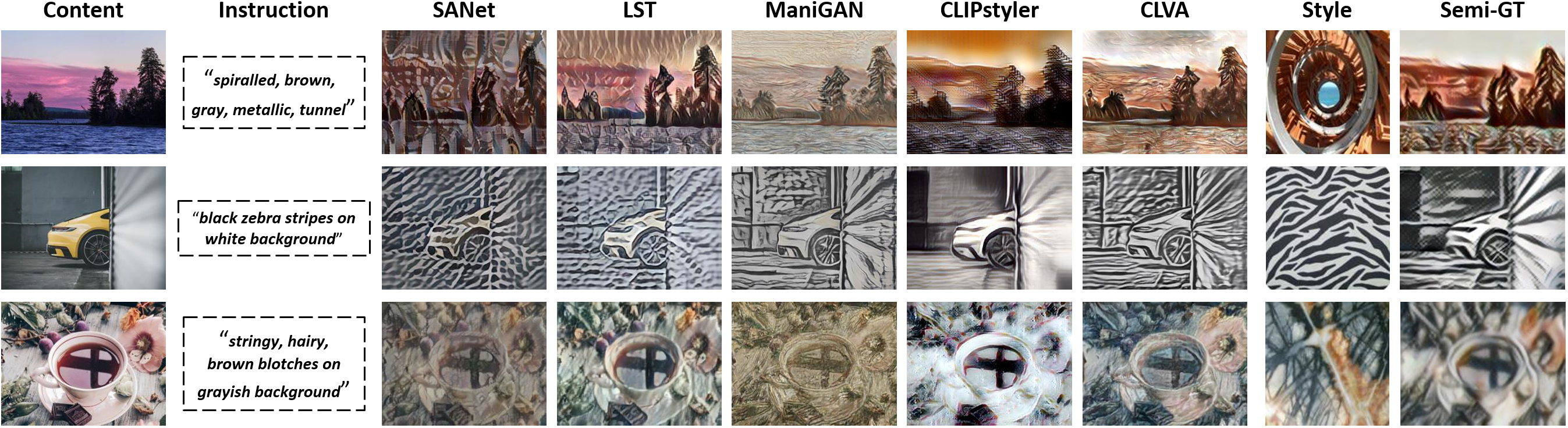}
    \vspace{-5ex}
    \caption{Visualized comparison using visual attribute instructions on DTD$^2$.}
    \vspace{-4ex}
    \label{fig:dtd}
\end{figure}

From the aspect of visualized comparison in Fig.~\ref{fig:dtd}, previous SANet and LST only produce repetitive and disorder textures in their transferred results. ManiGAN modifies the style directly over pixels, suffering from blurring objects; this deficiency can also be found in Table 1 (lower SSIM and lower Content matching). CLIPstyler is sometimes misguided by CLIP, making irrelevant patterns, such as the bright white background in the third case. Contrary to baselines, CLVA extracts a more detailed style from different kinds of guidance (\textit{``brown metallic''} in the first row and \textit{``stringy hairy''} in the third case), leading to superior \texttt{LDAST} results that correspond to style instructions.

\begin{table}[t]
\centering \scriptsize
    \begin{tabular}[t]{cccccccccc}
        \toprule
        ~ & \multicolumn{4}{c}{\textbf{Automatic Metrics}} & ~ & \multicolumn{4}{c}{\textbf{Human Evaluation}} \\
        \cmidrule{2-5} \cmidrule{7-10} 
        Method & SSIM$\uparrow$ & Percept$\downarrow$ & FAD$\downarrow$ & VLS$\uparrow$ & ~ & Content$\uparrow$ & Instruction$\uparrow$ & Style$\uparrow$ & Semi-GT$\uparrow$ \\
        \midrule
        SANet~\cite{park2019sa} & 38.36 & \textbf{0.0352} & \underline{0.1548} & 19.30 & ~ & \underline{3.170} & 2.978 & 2.980 & 2.890 \\
        LST~\cite{li2019lst} & \textbf{42.13} & 0.0386 & 0.1595 & 19.92 & ~ & 2.967 & 2.714 & 2.614 & 2.757 \\
        ManiGAN~\cite{li2020lbie} & 38.46 & 0.0500 & 0.1554 & 19.69 & ~ & 2.729 & 2.583 & 2.879 & \underline{3.192} \\
        CLIPstyler~\cite{kwon2022clip-styler} & 24.17 & 0.0659 & 0.1759 & \textbf{21.04} & ~ & 2.777 & \underline{3.140} & \underline{2.998} & 2.952 \\
        CLVA & \underline{40.32} & \underline{0.0357} & \textbf{0.1418} & \underline{20.11} & ~ & \textbf{3.357} & \textbf{3.586} & \textbf{3.530} & \textbf{3.208} \\
        \bottomrule
        \\
    \end{tabular}
    \vspace{-1ex}
    \caption{Testing results of \texttt{LDAST} using emotional effect instructions on ArtEmis.}
    \vspace{-4ex}
    \label{table:artemis}
\end{table}
\begin{figure}[!t]
\centering
    \includegraphics[width=\linewidth]{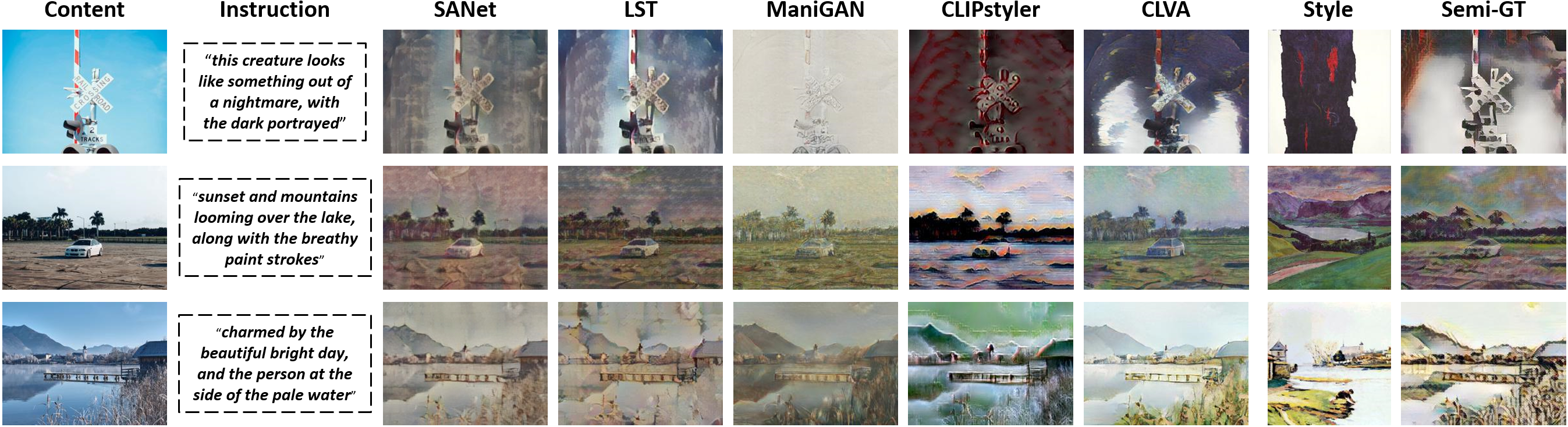}
    \vspace{-5ex}
    \caption{Visualized comparison using emotional effect instructions on ArtEmis.}
    \vspace{-4ex}
    \label{fig:artemis}
\end{figure}

\vspace{1ex} \noindent \textbf{Instruction with Emotional Effects.}
Unlike visual attributes, emotional effect instructions are more challenging as connecting to visual semantics of described objects or style patterns from human feelings. For example, ``\textit{yellowish and green}'' from ``\textit{sunset and mountains}'' or ``\textit{scaring charcoal grey}'' from ``\textit{nightmare}''. We consider this human style feeling on ArtEmis~\cite{achlioptas2021artemis}, where the model has to express the latent visual concepts of emotional effect instructions. CLVA performs with more balance (both second-highest SSIM and second-lowest Percept) from Table 2, especially the lowest 0.1418 FAD, making the most similar transferred results to Semi-GT. Though CLIPstyler~\cite{kwon2022clip-styler} achieves higher VLS by optimizating over CLIP, from human aspects, CLVA can preserve more concrete contents and present more correlated style patterns (higher 3.357 content and 3.586 instruction matching). The visualized comparison in Fig. 4 illustrates that previous SANet~\cite{park2019sa} and LST~\cite{li2019lst} contain unsmooth and fragmentary patterns with blurring contents. Without a style transformation process, ManiGAN~\cite{li2020lbie} modifies with only monotonous colors. CLIPstyler is failed to capture human style feelings well, suffering from weird and unpleasant results. Different from them, our CLVA learns the visual semantic during contrastive reasoning by comparing mutual relativeness between literal instructions and style images, leading to a more colorful and corresponding stylization as human emotion. More surprisingly, despite not instructed literally, CLVA perceives ``\textit{side of the water}'' and reveals the latent yet correlated ``\textit{grassland}'' precisely in the third row.

\begin{figure}[!t]
\centering
    \includegraphics[width=\linewidth]{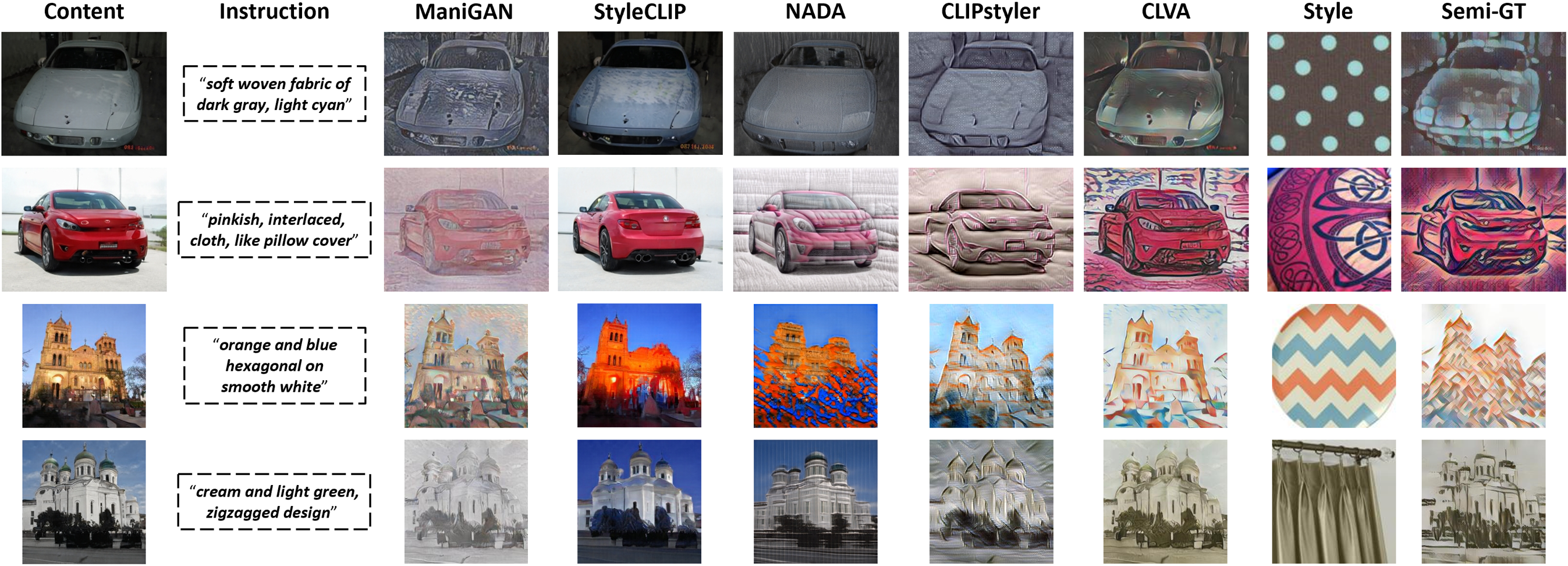}
    \vspace{-5ex}
    \caption{Visualized comparison on specific content domain (Car and Church).}
    \vspace{-4ex}
    \label{fig:domain}
\end{figure}

\vspace{1ex} \noindent \textbf{Specific Content Domain.}
To compare with StyleCLIP~\cite{patashnik2021style-clip} and NADA~\cite{gal2021nada} that are restricted by the pre-trained generator, we evaluate LDAST on the specific content domain. We consider the same domain images in StyleGAN2~\cite{kerras2020style-gan} and visual attribute instructions on DTD$^2$. Table~\ref{table:domain} indicates the numerical comparison on Car and Church. Our CLVA still produces superior results and is the most admirable by human. Since StyleCLIP and NADA rely on StyleGAN, they can only preserve content (highest 3.459 Content by StyleCLIP) but with limited stylization (lower Instruction and Style). Similar observations can be found in Fig.~\ref{fig:domain}, where StyleCLIP shows almost no modification for the second car. They can neither deal with the background; NADA even destroys the scene in the third row. In contrast to CLIPstyler~\cite{kwon2022clip-styler} that only contains abstractive and obscure styles, CLVA presents the detailed \textit{``read interplaced cloth''} behind the car and the color \textit{``cream''} precisely on the surface of the church.

\begin{table}[t]
\centering \scriptsize
    \begin{tabular}[t]{cccccccccc}
        \toprule
        ~ & \multicolumn{4}{c}{\textbf{Automatic Metrics}} & ~ & \multicolumn{4}{c}{\textbf{Human Evaluation}} \\
        \cmidrule{2-5} \cmidrule{7-10} 
        Method & SSIM$\uparrow$ & Percept$\downarrow$ & FAD$\downarrow$ & VLS$\uparrow$ & ~ & Content$\uparrow$ & Instruction$\uparrow$ & Style$\uparrow$ & Semi-GT$\uparrow$ \\
        \midrule
        ManiGAN~\cite{li2020lbie} & 26.45 & \underline{0.2329} & \underline{0.1672} & 23.44 & ~ & 2.861 & 2.894 & 2.978 & 2.893 \\
        StyleCLIP~\cite{patashnik2021style-clip} & \underline{28.03} & 0.2609 & 0.1812 & 21.55 & ~ & \textbf{3.459} & 2.845 & 2.930 & 2.829 \\
        NADA~\cite{gal2021nada} & 16.98 & 0.2733 & 0.1876 & 23.38 & ~ & 2.542 & 2.798 & 2.846 & 2.932 \\
        CLIPstyler~\cite{kwon2022clip-styler} & 18.43 & 0.2493 & 0.1826 & \textbf{24.16} & ~ & 2.986 & \underline{3.067} & \underline{3.003} & \underline{3.032} \\
        CLVA & \textbf{30.98} & \textbf{0.1957} & \textbf{0.1544} & \underline{23.68} & ~ & \underline{3.153} & \textbf{3.465} & \textbf{3.344} & \textbf{3.315} \\
        \bottomrule
        \\
    \end{tabular}
    \vspace{-1ex}
    \caption{Testing results of \texttt{LDAST} on specific content domain (Car and Church).}
    \vspace{-4ex}
    \label{table:domain}
\end{table}

\begin{table}[t]
\centering \scriptsize
    \begin{tabular}[t]{ccccccccccc}
        \toprule
        ~ & \multicolumn{4}{c}{\textbf{Ablation Settings}} & ~ & \multicolumn{4}{c}{\textbf{Automatic Metrics}} \\
        \cmidrule{2-5} \cmidrule{7-10} 
        ~ & $\mathcal{L}_\text{rec}$+$\mathcal{L}_\text{psd}$ & $\mathcal{L}_\text{cm}$ & $\mathcal{L}_\text{sm}$ & $\mathcal{L}_\text{ctr}$ & ~ & SSIM$\uparrow$ & Percept$\downarrow$ & FAD$\downarrow$ & VLS$\uparrow$ \\
        \midrule
        (a) & \ding{51} & \ding{55} & \ding{55} & \ding{55} & ~ & 34.73 & 0.2290 & 0.1568 & 23.29 \\
        (b) & \ding{51} & \ding{51} & \ding{55} & \ding{55} & ~ & \underline{36.05} & 0.2304 & 0.1512 & 23.27 \\
        (c) & \ding{51} & \ding{55} & \ding{51} & \ding{55} & ~ & 35.73 & \underline{0.2049} & \underline{0.1508} & \underline{23.69} \\
        (d) & \ding{51} & \ding{51} & \ding{51} & \ding{55} & & 35.86 & 0.2100 & 0.1499 & 23.54 \\
        (e) & \ding{51} & \ding{51} & \ding{51} & \ding{51} & ~ & \textbf{36.65} & \textbf{0.2033} & \textbf{0.1493} & \textbf{24.00} \\
        \bottomrule
        \\
    \end{tabular}
    \vspace{-1ex}
    \caption{Ablation study of CLVA using visual attribute instructions on DTD$^2$.}
    \vspace{-4ex}
    \label{table:abl}
\end{table}

\subsection{Ablation Study}
\vspace{-0.5ex} We conduct an ablation study of each component effect on DTD$^2$ in Table~\ref{table:abl}. At row (a), with the reconstruction $\mathcal{L}_\text{rec}$ and the patch-wise style $\mathcal{L}_\text{psd}$, CLVA achieves feasible \texttt{LDAST} results by concrete structures and extracted style semantics. Row (b)-(d) shows the strength of content matching $\mathcal{L}_\text{cm}$ and style matching $\mathcal{L}_\text{sm}$. In particular, content matching helps the structure similarity to content images (higher 36.05 SSIM). Style matching aims at analogous visual patterns to style images, which leads to better stylization quality (lower 0.2049 Percept and higher 23.69 VLS). If considering altogether, it can benefit and strike a balance between both. Finally, contrastive reasoning $\mathcal{L}_\text{ctr}$ further enables CLVA to consider contrastive pairs, making a comprehensive improvement at row (e).

\begin{figure}[t]
\centering
\begin{minipage}{.38\linewidth}
\centering \scriptsize
    \begin{tabular}[t]{cccccc}
        \toprule
        ~ & \multicolumn{2}{c}{\textbf{DTD$^2$}} & ~ & \multicolumn{2}{c}{\textbf{ArtEmis}} \\
        \cmidrule{2-3} \cmidrule{5-6}
        Method & R@1 & R@5 & ~ & R@1 & R@5 \\
        \midrule
        CLIP~\cite{shi2020clip} & 13.9 & 30.7 & ~ & 9.8 & 20.7 \\
        CLVA & \textbf{19.3} & \textbf{45.1} & ~ & \textbf{13.9} & \textbf{30.7} \\
        \bottomrule
        \\
    \end{tabular}
    \vspace{-4ex}
    \captionof{table}{Instruction-to-style retrieval on DTD$^2$ and ArtEmis.}
    \label{table:i2s}
\end{minipage}~~
\begin{minipage}{.60\linewidth}
\centering \scriptsize
    \begin{tabular}[t]{ccccc}
        \toprule
        ~ & \multicolumn{4}{c}{\textbf{Human Evaluation}} \\
        \cmidrule{2-5}
        Method & Content$\uparrow$ & Instruction$\uparrow$ & Style$\uparrow$ & Semi-GT$\uparrow$ \\
        \midrule
        CLIPstyler (ft.) & 1.208 & 1.347 & 1.292 & 1.333 \\
        CLVA & \textbf{1.792} & \textbf{1.653} & \textbf{1.708} & \textbf{1.667} \\
        \bottomrule
        \\
    \end{tabular}
    \vspace{-4ex}
    \captionof{table}{Human comparison between CLVA and CLIPstyle with fine-tuned CLIP on DTD$^2$.}
    \label{table:finetune}
\end{minipage}
\end{figure}

\begin{figure}[!t]
\centering
    \includegraphics[width=\linewidth]{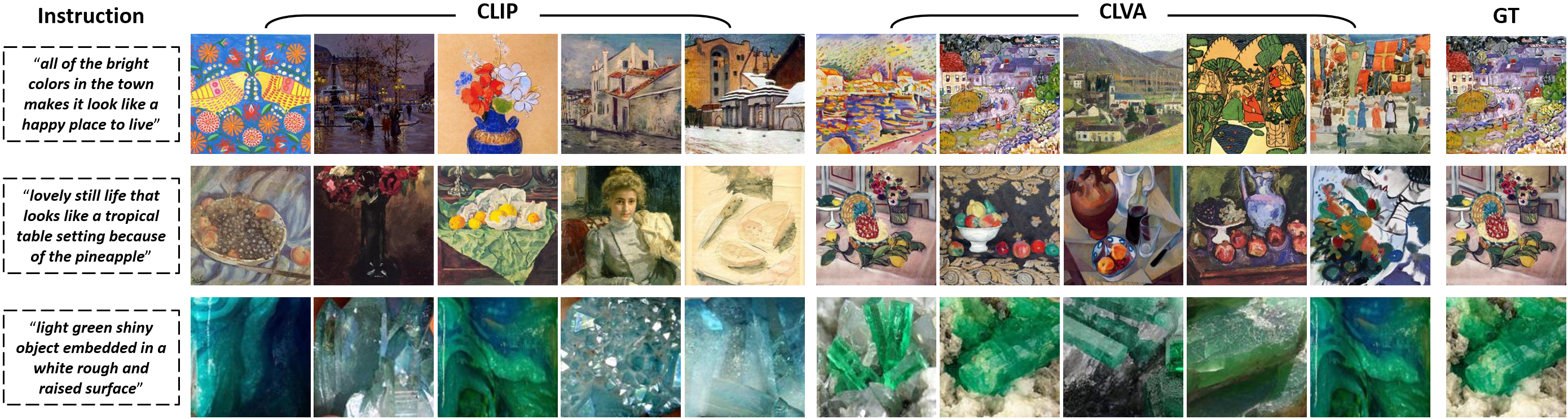}
    \vspace{-5ex}
    \caption{Visualization examples of instruction-to-style retrieval by CLIP and CLVA.}
    \vspace{-4ex}
    \label{fig:i2s}
\end{figure}

\vspace{1ex} \noindent \textbf{Why CLVA is better than CLIP-based?}
Despite no CLIP optimized, CLVA demonstrates superior results on \texttt{LDAST} with all aspects of automatic metrics and human evaluation. To investigate it, we conduct instruction-to-style retrieval based on the similarity between features of style instructions and style images. Table~\ref{table:i2s} shows that our learned CLVA performs higher Recall@k on both DTD$^2$ and ArtEmis, leading to a better instruction-style alignment than the used CLIP. The visualization in Fig.~\ref{fig:i2s} also indicates the flaw of CLIP on detailed style patterns. For example, in the first row, CLIP only presents either \textit{``bright color''} or \textit{``town''}  in the retrieval results. In contrast, CLVA can capture both and present more related \texttt{LDAST} to \textit{``happy place to live''}. From Table~\ref{table:finetune}, even CLIP has been fine-tuned ahead; our CLVA still produces preferrable \texttt{LDAST} results from all human aspects of content, instruction, and style matching. This observation supports that contrastive reasoning, which considers contrastive pairs of content images and style instructions, is required to benefit from mutual relativeness. 

Apart from transfer quality, CLVA also holds a higher efficiency than CLIP-based methods. Table~\ref{table:eff} illustrates the time and GPU cost on a single TITAN X (12GB) with content image size 256x192. All CLIP-based methods take more than 30 seconds for only one pair of content images and style instructions. Instead of numerous iterations to align with CLIP, we extract style semantics and carry out \texttt{LDAST} in one shot, taking merely 0.03 seconds for one input. Without updating the model during inference, CLVA supports parallelization and can accomplish 50 pairs in half a second. Besides, as a lightweight style transfer network, CLVA requires the least GPU memory for \texttt{LDAST}. In summary, our CLVA surpasses those CLIP-based methods on both quality and efficiency because of the detailed style deficiency and the required optimizing iteration from CLIP.

\begin{table}[t]
\centering \scriptsize
    \begin{tabular}[t]{cccccccc}
        \toprule
        ~ & \multicolumn{3}{c}{\textbf{Time (sec)}} & ~ & \multicolumn{3}{c}{\textbf{GPU (MB)}} \\
        \cmidrule{2-4} \cmidrule{6-8}
        Method & BS=1 & 32 & 50 & ~ & BS=1 & 32 & 50 \\
        \midrule
        ManiGAN~\cite{li2020lbie} & 0.079 & 0.533 & 1.148 & ~ & 3312 & 6572 & 8129 \\
        StyleCLIP~\cite{patashnik2021style-clip} & 32.38 & * & * & ~ & 4149 & * & * \\
        NADA~\cite{gal2021nada} & 63.49 & * & * & ~ & 6413 & * & * \\
        CLIPstyler~\cite{kwon2022clip-styler} & 99.98 & * & * & ~ & 5429 & * & * \\
        CLVA & \textbf{0.029} & \textbf{0.246} & \textbf{0.405} & ~ & \textbf{1525} & \textbf{3207} & \textbf{4441} \\
        \bottomrule
        \\
    \end{tabular}
    \vspace{-1ex}
    \caption{Time and GPU cost when performing \texttt{LDAST} on TITAN X with content image size 256x192. * means this method can only run one input at a time.}
    \vspace{-4ex}
    \label{table:eff}
\end{table}

\begin{figure}[!t]
\centering
    \includegraphics[width=\linewidth]{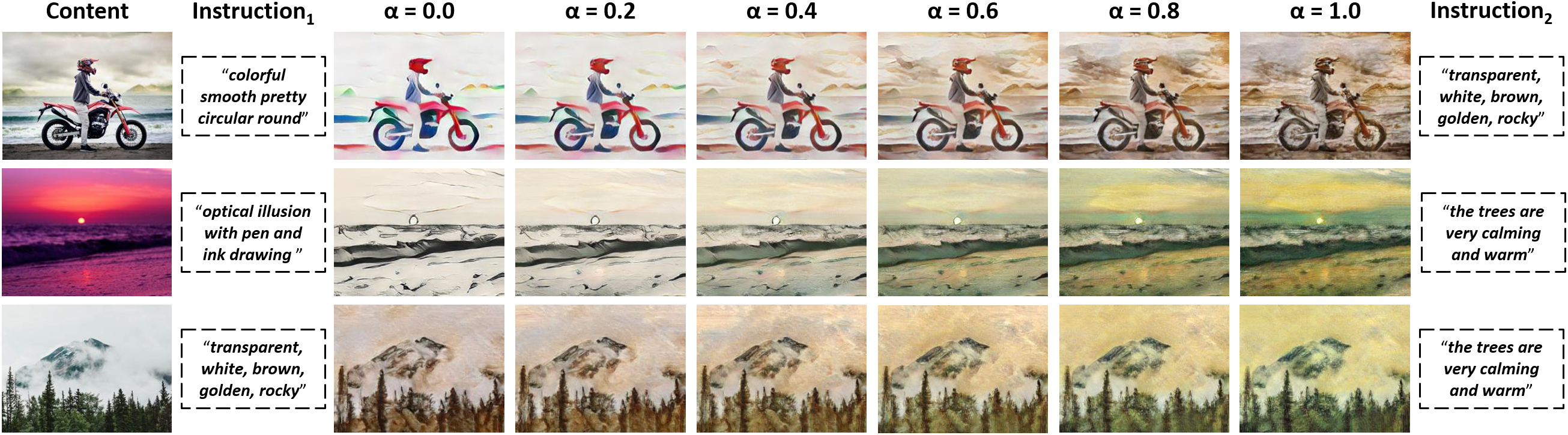}
    \vspace{-5ex}
    \caption{Style interpolation results of \texttt{LDAST} over instructions.}
    \vspace{-4ex}
    \label{fig:interpolation}
\end{figure}

\begin{figure}[!t]
\centering
    \includegraphics[width=\linewidth]{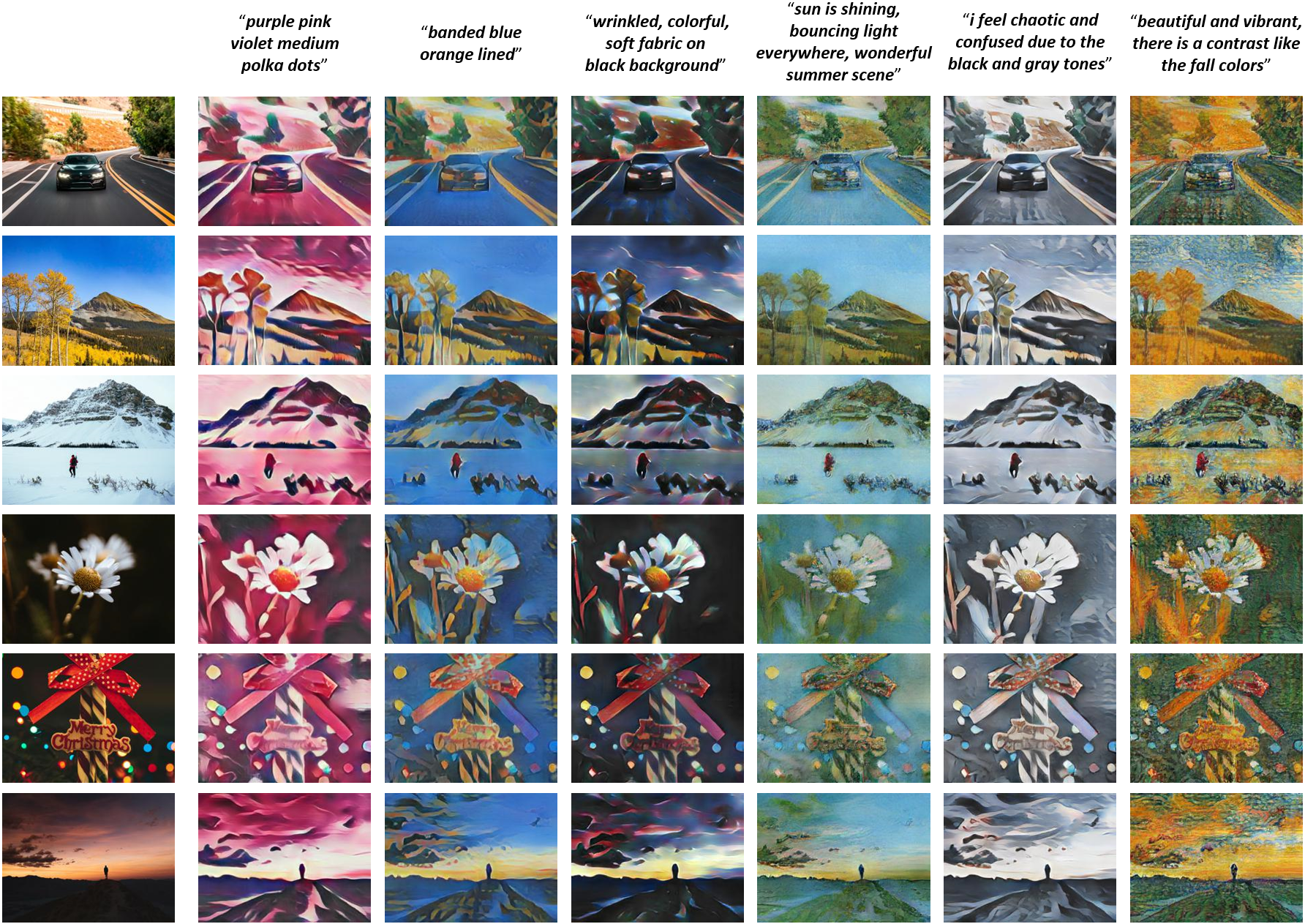}
    \vspace{-5ex}
    \caption{CLVA results on diverse pairs of content images and style instructions.}
    \vspace{-3ex}
    \label{fig:qual}
\end{figure}

\begin{figure}[!t]
\centering
    \includegraphics[width=\linewidth]{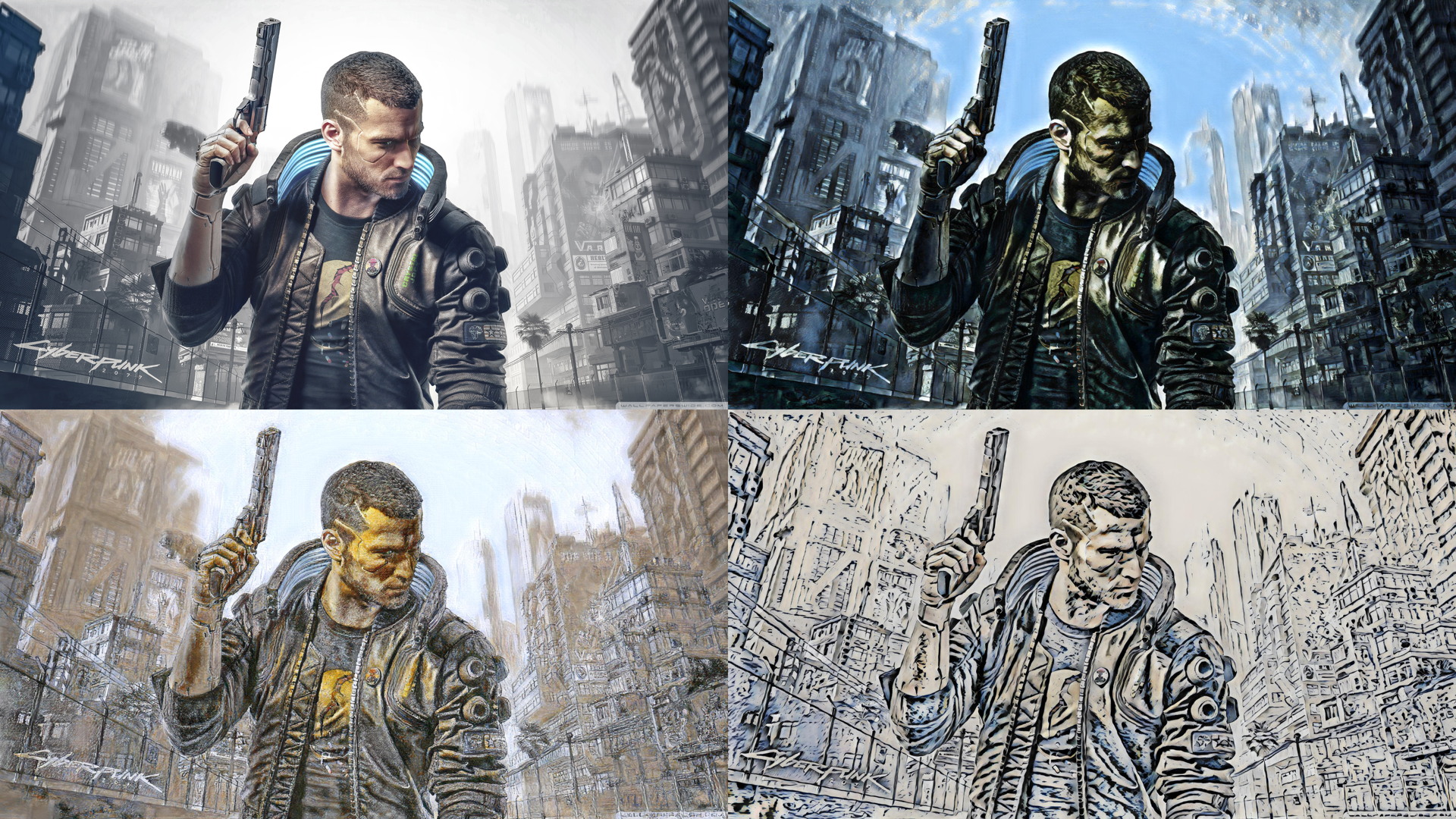} % 2213, 2213, 37254
    \vspace{-5ex}
    \caption{High-resolution (1920x1080) \texttt{LDAST} results by CLVA with upper right: \textit{``the lonely world makes me feel scared and nostalgic how sky and sea merge together''}; lower left: \textit{``the snow and lights in the shop windows looks like a winter scene''}; lower right: \textit{``ink painting, black dotted line, whiteboard''}.}
    \vspace{-3ex}
    \label{fig:high}
\end{figure}

\vspace{1ex} \noindent \textbf{Qualitative Results.}
As shown in Fig.~\ref{fig:interpolation}, we investigate the linear interpolation of extracted style patterns by CLVA. Considering style features $h^\mathcal{S}_{\mathcal{X}_1}$ and $h^\mathcal{S}_{\mathcal{X}_2}$  of instructions $\mathcal{X}_\text{1}$ and $\mathcal{X}_\text{2}$, the interpolated $h^\mathcal{S}_\text{p}$ should be:
\begin{equation}
\begin{split}
    h^\mathcal{S}_\text{p} &= (1-\alpha) h^\mathcal{S}_{\mathcal{X}_1} + \alpha h^\mathcal{S}_{\mathcal{X}_2},
\end{split}
\end{equation}
where $\alpha$ is the style ratio between the two. Fig.~\ref{fig:interpolation} presents a smooth transformation from one style instruction to another. By training on DTD$^2$ and ArtEmis altogether, CLVA even performs interpolated stylization by both visual attribute and emotional effect instructions in the third row. Fig.~\ref{fig:qual} illustrates diverse \texttt{LDAST} results by our CLVA. Since CLVA supports arbitrary content images, we can also modify the style detail for high-resolution inputs in Fig.~\ref{fig:high}.

\section{Conclusion}
We introduce language-driven artistic style transfer (\texttt{LDAST}) to do stylization for a content image by a style instruction. We propose contrastive language visual artist (CLVA) that adopts the patch-wise style discriminator and contrastive reasoning to jointly learn between style images and style instructions. We demonstrate that CLVA can express various style patterns of visual attributes as well as emotional effects and perform \texttt{LDAST} efficiently. CLVA also outperforms baselines on both automatic metrics and human evaluation. We believe that \texttt{LDAST} can make visual applications like image/video effect more controllable for humans.

\vspace{1ex} \noindent \textbf{Acknowledgments.}
Research was sponsored by the U.S. Army Research Office and was accomplished under Contract Number W911NF-19-D-0001 for the Institute for Collaborative Biotechnologies. The views and conclusions contained in this document are those of the authors and should not be interpreted as representing the official policies, either expressed or implied, of the U.S. Government. The U.S. Government is authorized to reproduce and distribute reprints for Government purposes notwithstanding
any copyright notation herein.

\clearpage

\clearpage

\appendix
\section{Implementation Detail}
We adopt VGG-19~\cite{simonyan2015vgg,park2019sa} as our visual encoder $G_\text{E}$ and visual decoder $G_\text{D}$. Text encoder $\phi$ first adopts RoBERTa~\cite{liu2019roberta,reimers2019roberta} for a general linguistic and then expands its spatial dimension to jointly embed with style features. We follow the self-attention layer from SANet~\cite{park2019sa} to fuse between content and style features in $G_\text{D}$. The patch-wise style discriminator $D$ contains a similar architecture with a dense layer to determine the correlation between instructions and image patches. Both $G_\text{E}$ and $G_\text{D}$ are initialized from SANet and further update during the CLVA training process. We adopt Adam~\cite{kingma2015adam} to optimize CLVA with learning rate 3e-4 for $\mathcal{L}_G$, 1e-4 for $\mathcal{L}_D$, and 3e-5 for $\mathcal{L}_\text{ctr}$.

\section{Complete Results on all Baselines}
Table~\ref{table:complete-dtd} shows the complete \texttt{LDAST} results using visual attribute instructions on DTD$^2$~\cite{cimpoi14dtd}. As with previous style transfer methods, NST~\cite{gatys2016nst} and WCT~\cite{li2017wct} cannot handle the target style well (higher Percept loss), resulting in the irrelevant produced patterns to style instructions (lower VLS). AdaIn~\cite{huang2017adain} better performs stylization from guided texts, but the content will be much modified with a relatively lower SSIM. Similar observations can be found in Fig.~\ref{fig:complete-dtd}, where the scissors and color pencils by AdaIn are mostly distorted in the second row. In contrast, CLVA presents the detailed multi-color pattern (the second case) yet preserves the concrete structure of the flower (the third case) at the same time.

We also illustrate the results of SANet~\cite{park2019sa} and LST~\cite{li2019lst} on specific content domain (Car and Church) in Table~\ref{table:complete-domain}. By learning from pairs of style instructions and images, style transfer methods can perform better stylization than StyleGAN-restricted methods (StyleCLIP~\cite{patashnik2021style-clip} and NADA~\cite{gal2021nada}). Despite having similar results on automatic metrics, there is a noticeable quality gap compared to our CLVA in Fig.~\ref{fig:complete-domain}. Unlike SANet and LST, which contain repetitive and chaotic patterns, CLVA accomplishes \texttt{LDAST} with even more concrete contents (the second car and the fourth church) than Semi-GT through consistent matching during contrastive reasoning.

\vspace{-2ex}
\begin{figure}[!ht]
\centering
\begin{minipage}{.49\linewidth}
\centering \scriptsize
    \begin{tabular}[t]{ccccc}
        \toprule
        ~ & \multicolumn{4}{c}{\textbf{Automatic Metrics}} \\
        \cmidrule{2-5}
        Method & SSIM$\uparrow$ & Percept$\downarrow$ & FAD$\downarrow$ & VLS$\uparrow$ \\
        \midrule
        NST~\cite{gatys2016nst} & 19.70 & 0.2441 & 0.1920 & 18.97 \\
        WCT~\cite{li2017wct} & \textbf{37.88} & 0.2617 & 0.1720 & 19.76 \\
        AdaIn~\cite{huang2017adain} & 29.43 & \underline{0.2081} & 0.1748 & 21.65 \\
        SANet~\cite{park2019sa} & 35.50 & 0.2129 & 0.1627 & 23.57 \\
        LST~\cite{li2019lst} & 34.84 & 0.2137 & \underline{0.1533} & 23.16 \\
        ManiGAN~\cite{li2020lbie} & 32.7 & 0.2401 & 0.1663 & 23.25 \\
        CLIPstyler~\cite{kwon2022clip-styler} & 25.24 & 0.2598 & 0.1818 & \textbf{24.62} \\
        CLVA & \underline{36.65} & \textbf{0.2033} & \textbf{0.1493} & \underline{24.00} \\
        \bottomrule
        \\
    \end{tabular}
    \vspace{-4ex}
    \captionof{table}{Complete results using visual attribute instructions on DTD$^2$.}
    \label{table:complete-dtd}
\end{minipage}~~
\begin{minipage}{.49\linewidth}
\centering \scriptsize
    \begin{tabular}[t]{ccccc}
        \toprule
        ~ & \multicolumn{4}{c}{\textbf{Automatic Metrics}} \\
        \cmidrule{2-5}
        Method & SSIM$\uparrow$ & Percept$\downarrow$ & FAD$\downarrow$ & VLS$\uparrow$ \\
        \midrule
        SANet~\cite{park2019sa} & 30.95 & \underline{0.1982} & 0.1638 & 23.20 \\
        LST~\cite{li2019lst} & \textbf{31.16} & 0.2045 & \underline{0.1606} & 23.34 \\
        ManiGAN~\cite{li2020lbie} & 26.45 & 0.2329 & 0.1672 & 23.44 \\
        StyleCLIP~\cite{patashnik2021style-clip} & 28.03 & 0.2609 & 0.1812 & 21.55 \\
        NADA~\cite{gal2021nada} & 16.98 & 0.2733 & 0.1876 & 23.38 \\
        CLIPstyler~\cite{kwon2022clip-styler} & 18.43 & 0.2493 & 0.1826 & \textbf{24.16} \\
        CLVA & \underline{30.98} & \textbf{0.1957} & \textbf{0.1544} & \underline{23.68} \\
        \bottomrule
        \\
    \end{tabular}
    \vspace{-4ex}
    \captionof{table}{Complete results on specific content domain (Car and Church).}
    \label{table:complete-domain}
\end{minipage}
\end{figure}

\begin{figure}[!t]
\centering
    \includegraphics[width=\linewidth]{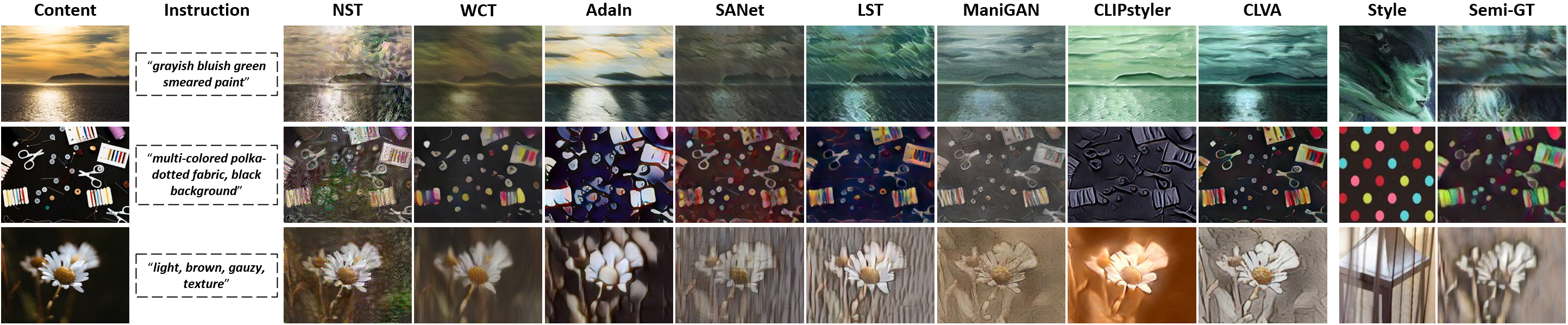}
    \vspace{-5ex}
    \caption{Complete visualization using visual attribute instructions on DTD$^2$.}
    \label{fig:complete-dtd}
\end{figure}

\begin{figure}[!t]
\centering
    \includegraphics[width=.92\linewidth]{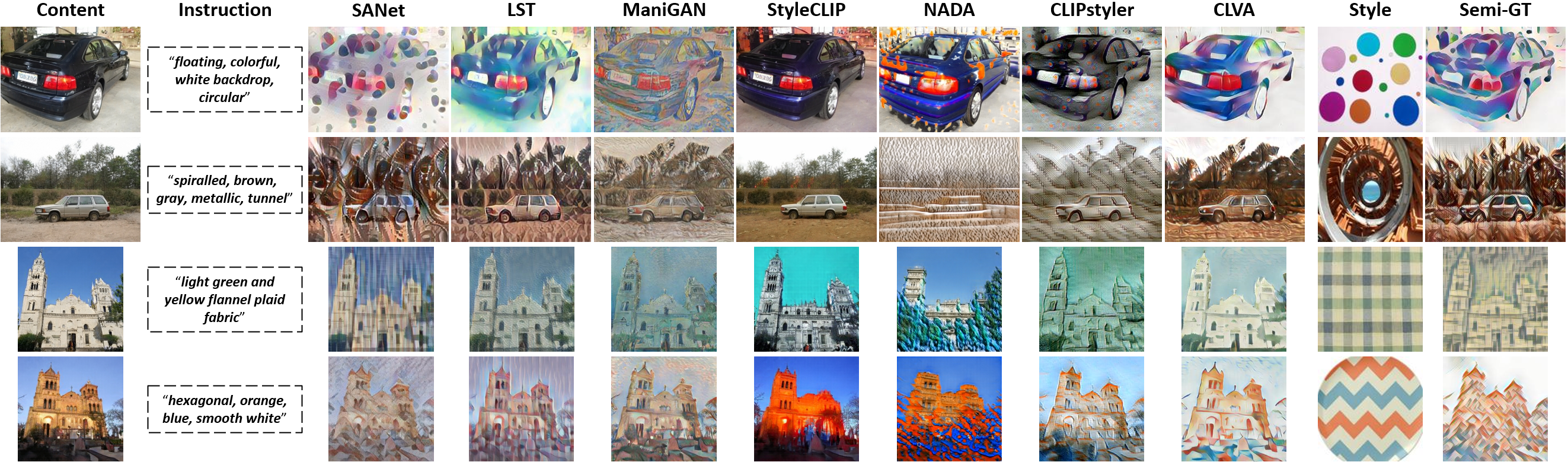}
    \vspace{-1ex}
    \caption{Complete visualization on specific content domain (Car and Church).}
    \vspace{-1ex}
    \label{fig:complete-domain}
\end{figure}

\section{Retrieval-based Baseline}
Apart from producing the transferred result by the style instruction $\mathcal{X}$, we also investigate a two-step retrieval-based baseline. We first adopt the CLIP~\cite{shi2020clip} alignment to find the most similar style image $\mathcal{S}$ (with 13.9 R@1 and 30.7 R@5 on DTD$^2$~\cite{cimpoi14dtd}) via $\mathcal{X}$. Then, the retrieved $\mathcal{S}$ is used to carry out standard style transfer. Table~\ref{table:rtrv} shows that the two-step retrieval baseline performs slightly better on visual similarity to Semi-GT. However, by stylization from guided texts, CLVA produces more correlated style patterns to instructions (higher 24.00 VLS). In addition, this retrieval-based method still relies on an existing set of style images and limits the diversity of stylization due to the collection size.

\vspace{-2ex}
\begin{table}[!ht]
\centering \scriptsize
    \begin{tabular}[t]{ccccc}
        \toprule
        ~ & \multicolumn{4}{c}{\textbf{Automatic Metrics}} \\
        \cmidrule{2-5}
        Method & SSIM$\uparrow$ & Percept$\downarrow$ & FAD$\downarrow$ & VLS$\uparrow$ \\
        \midrule
        SANet~\cite{park2019sa} & 35.50 & 0.2129 & 0.1627 & 23.57 \\
        SANet (rtrv.) & \textbf{37.74} & \textbf{0.2005} & \textbf{0.1421} & \underline{23.68} \\
        CLVA & \underline{36.65} & \underline{0.2033} & \underline{0.1493} & \textbf{24.00} \\
        \bottomrule
        \\
    \end{tabular}
    \caption{Testing results of the two-step retrieval-based baseline using visual attribute instructions on DTD$^2$.}
    \vspace{-8ex}
    \label{table:rtrv}
\end{table}

\vspace{-2ex}
\section{Human Evaluation}
\begin{figure}[!ht]
\centering
    \includegraphics[width=\linewidth]{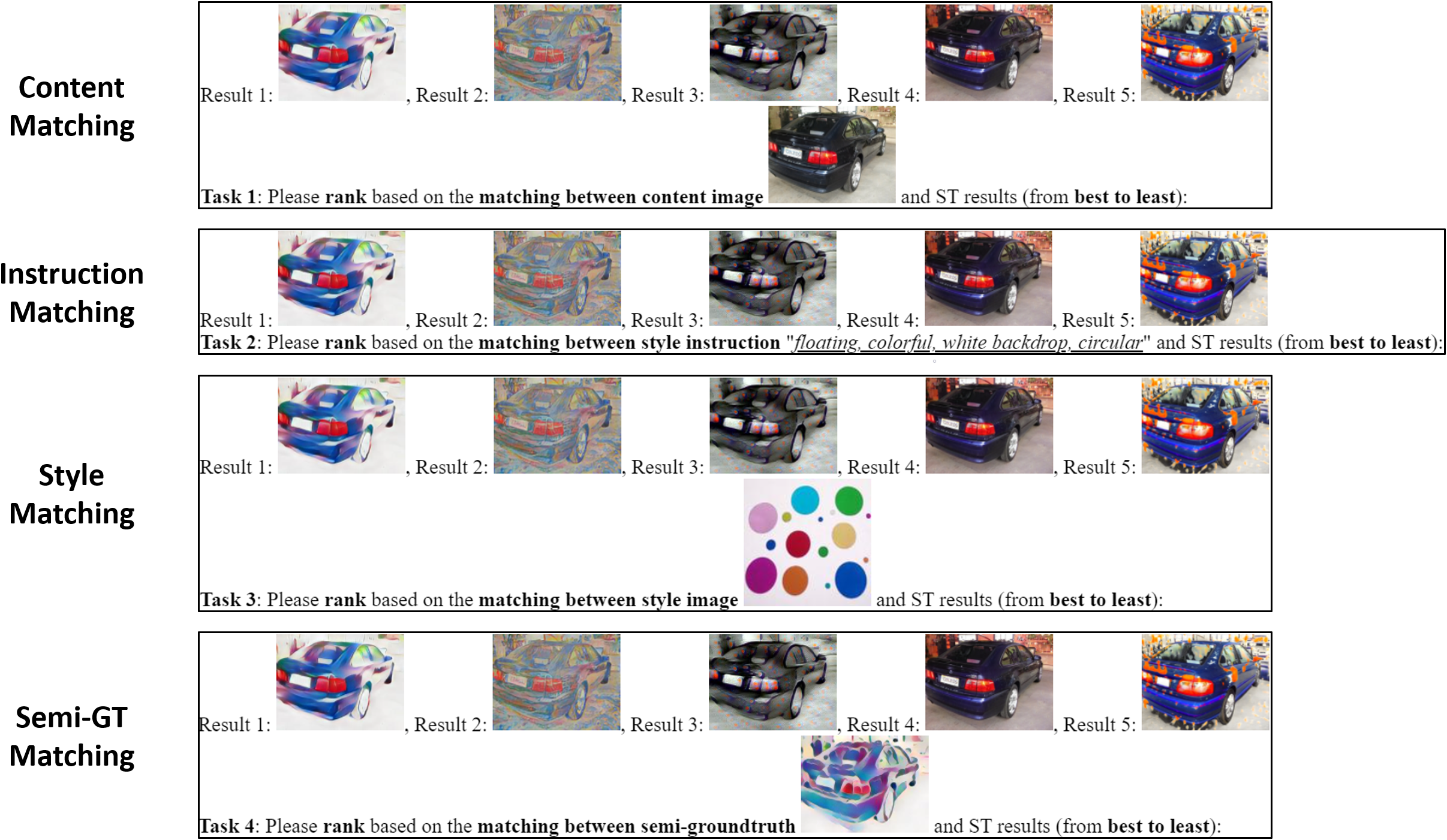}
    \vspace{-4ex}
    \caption{The screenshots of the ranking tasks for human evaluation on \texttt{LDAST}.}
    \vspace{-2ex}
    \label{fig:human}
\end{figure}

We investigate the quality of \texttt{LDAST} results from the human aspect through Amazon Mechanical Turk. Fig.~\ref{fig:human} illustrates the screenshots of the human tasks. MTurkers rank the correlation of the \texttt{LDAST} result according to Content, Instruction, Style, and Semi-GT matching. Each MTurker rewards \$2.0 and takes a mean of 15 minutes.

\vspace{-1ex}

\section{Limitation and Ethics Discussion}
Though our work benefits creative visual applications, there are several remaining technical issues. At first, complicated instructions that contain excessive visual attributes or emotional effects are still difficult to address by our CLVA. CLVA may lean towards specific visual concepts, resulting in correlated but monotonous stylization. Secondly, since the learning of CLVA relies on patch-wise style discriminator $D$, the quality of the randomly sampled patches will crucially influence the transferred results. On the other hand, there may be a ``fake as real'' doubt for those manipulated content images. To mitigate this issue, we can apply techniques from image forensics \cite{wang2020ethic,huh2018ethic,frank2020ethic} to detect the authenticity of an image. Regarding guided instructions, for example, hate speech detection \cite{aluru2020ethic,huang2020ethic,samghabadi2020ethic,samanta2019ethic} can help to filter out malicious texts and prevent from producing controversial results with ethics concerns.

\bibliographystyle{splncs04}
\bibliography{egbib}

\end{document}